\documentclass[letterpaper, 10 pt, conference]{ieeeconf}  

\IEEEoverridecommandlockouts                              

\usepackage{url}
\usepackage{soul}
\usepackage{graphicx}
\usepackage{wrapfig}
\usepackage{amsmath}
\usepackage{amssymb}
\usepackage{booktabs}
\usepackage{xcolor}
\usepackage{multirow}

\makeatletter
\let\labelindent\undefined
\makeatother
\usepackage{mdframed}
\usepackage{enumitem}
\usepackage{setspace}
\usepackage[acronym]{glossaries}
\usepackage{glossaries-extra}
\usepackage{cite}
\usepackage[hidelinks]{hyperref}
\usepackage[capitalise]{cleveref}

\glsdisablehyper
\setabbreviationstyle[acronym]{long-short}
\setabbreviationstyle[short]{short-nolong}
\newacronym[category={short}]{GPT}{\textsc{gpt}}{GPT}
\newacronym[category={short}]{GPT-OSS}{\textsc{gpt-oss}}{GPT-OSS}
\newacronym{LLM}{\textsc{llm}}{large language model}
\newacronym{VLM}{\textsc{vlm}}{vision language model}
\newacronym[category={short}]{ProcTHOR}{\textsc{p}roc\textsc{thor}}{\textsc{p}roc\textsc{thor}}
\newacronym{LSP}{\textsc{lsp}}{learning over subgoals planning}
\newacronym{UCB}{\textsc{ucb}}{upper confidence bound}
\newacronym{PDDL}{\textsc{pddl}}{planning domain definition language}
\newacronym{API}{\textsc{api}}{application programming interface}

\newcommand{\ourplanner}{\textsc{llm+model}}
\newcommand{\fulllm}{\textsc{llm-direct}}
\newcommand{\optimistic}{\textsc{optimistic+greedy}}
\newcommand{\prompta}{\textsc{p-context-a}}
\newcommand{\promptb}{\textsc{p-context-b}}
\newcommand{\promptc}{\textsc{p-minimal}}
\newcommand{\promptd}{\textsc{p-direct}}

\newcommand{\citet}{\cite}
\newcommand{\citep}{\cite}

\definecolor{tabblue}{rgb}{0.12156863, 0.46666667, 0.70588235}
\definecolor{taborange}{rgb}{1.0, 0.49803922, 0.05490196}
\definecolor{tabgreen}{rgb}{0.17254902, 0.62745098, 0.17254902}
\DeclareMathOperator*{\argmin}{argmin}
\title{\LARGE \bf
Object Search in Partially-Known Environments via LLM-informed Model-based Planning and Prompt Selection
}

\author{Abhishek Paudel$^{*}$, Abhish Khanal$^{*}$, Raihan I. Arnob, Shahriar Hossain, and Gregory J. Stein
\thanks{* denotes equal contribution. All authors are with Department of Computer Science at George Mason University, Fairfax, Virginia, USA. Corresponding author: Abhishek Paudel {\tt\small (apaudel4@gmu.edu)}
}
}

\begin{document}
\maketitle

\begin{abstract}
We present a novel \glsxtrshort{LLM}-informed model-based planning framework, and a novel prompt selection method, for object search  in partially-known environments. Our approach uses an \glsxtrshort{LLM} to estimate statistics about the likelihood of finding the target object when searching various locations throughout the scene that, combined with travel costs extracted from the environment map, are used to instantiate a model, thus using the \glsxtrshort{LLM} to inform planning and achieve effective search performance. Moreover, the abstraction upon which our approach relies is amenable to deployment-time model selection via the recent \emph{offline replay} approach, an insight we leverage to enable fast prompt and \glsxtrshort{LLM} selection during deployment. Simulation experiments demonstrate that our \glsxtrshort{LLM}-informed model-based planning approach outperforms the baseline planning strategy that fully relies on \glsxtrshort{LLM} and optimistic strategy with as much as 11.8\% and 39.2\% improvements respectively, and our bandit-like selection approach enables quick selection of best prompts and \glsxtrshortpl{LLM} resulting in 6.5\% lower average cost and 33.8\% lower average cumulative regret over baseline \glsxtrshort{UCB} bandit selection. Real-robot experiments in an apartment demonstrate similar improvements and so further validate our approach.
\end{abstract}

\section{Introduction}

We consider the problem of \emph{object search} in partially-known household environments, in which a robot is tasked to find an object of interest and can use \glspl{LLM} to inform the robot's behavior.
Effective object search in these scenarios often requires (i) considering the impacts of the robot's immediate actions far into the future and (ii) an ability for the robot to continuously self-evaluate during deployment to improve itself over time.

Model-based planning with a high-level action abstraction is a common decision-making framework for embodied intelligence, since it not only enables deciding what action the robot should do next but also enables reasoning farther into the future about the long-term value of a particular course of action.
While many existing approaches that leverage \glspl{LLM} for object search tasks use a high-level action abstraction in which the robot's actions corresponds to exploration of spaces that might contain the target object~\citep{dorbala2023can,zhou2023esc,yu2023l3mvn,arjun2024cognitive,ge2024commonsense,rajvanshi2024saynav}, they often prompt the \glspl{LLM} to directly decide what action the robot should pick next~\citep{zhou2023esc,dorbala2023can,yu2023l3mvn} without using a plannning framework.
As such, \glspl{LLM} can struggle to achieve good performance on many planning tasks~\citep{valmeekam2022large,valmeekam2023planning,kambhampati2024can,kambhampati2024llms}, particularly those necessary for embodied intelligence, including object search.
Moreover, it is well-established that \glspl{LLM} perform poorly on quantitative reasoning tasks~\citep{lewkowycz2022solving,arora2023have,boye2025large,rahman2025fragile}, a limitation that translates to difficulties in choosing the \emph{best performing} of a family of potentially suitable actions, resulting in greedy or myopic behavior that model-based reasoning is designed to avoid.
However, it is not straightforward to integrate an \gls{LLM} with a model-based planner and thus unclear how to best benefit from both model-based reasoning and the useful commonsense understanding an \gls{LLM} can provide.

\begin{figure*}
    \centering
    \vspace{0.6em}
    \includegraphics[width=\textwidth]{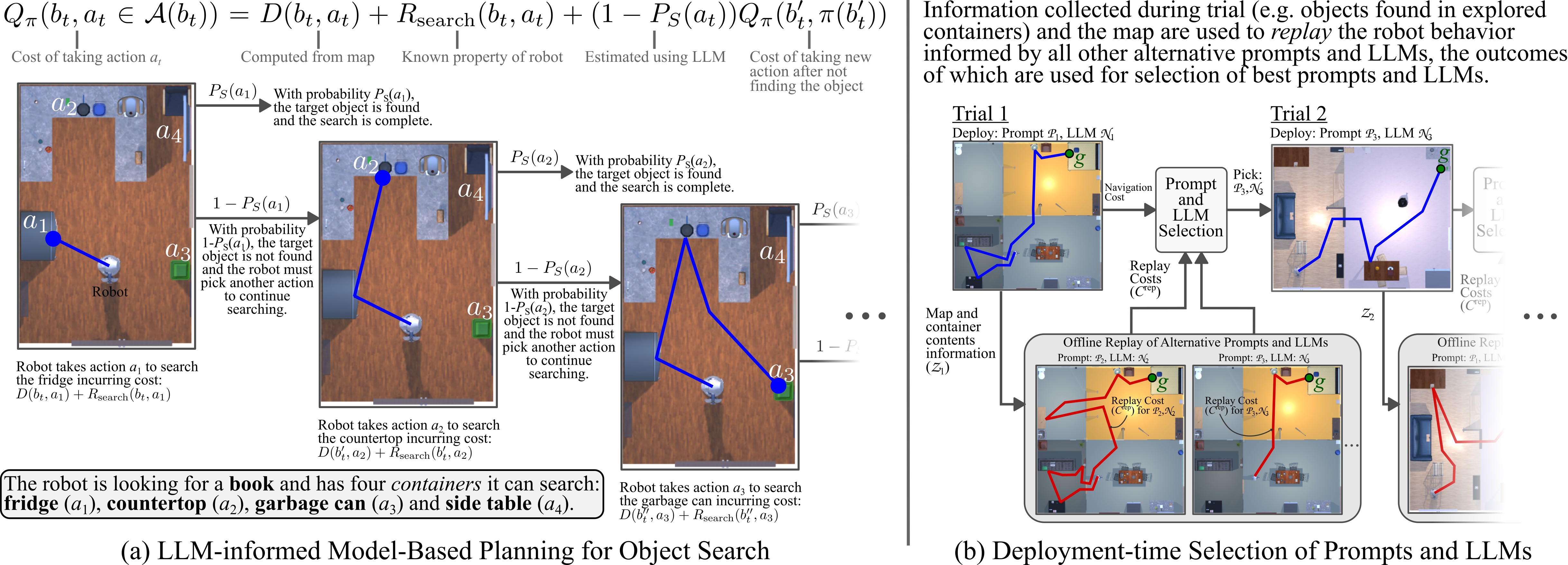}
    \vspace{-1.6em}
    \caption{Our \textbf{high-level action abstraction}, in which actions correspond to searching available containers to look for target object, enables (a) \textbf{model-based planning} in which an \textbf{\gls{LLM} informs the likelihood $P_S$ of finding target object} in a container and, (b) \textbf{fast deployment-time selection of prompts and \glspl{LLM}} for effective object search performance.}
    \vspace{-1.6em}
    \label{fig:overview}
\end{figure*}

In addition, the performance in object search tasks depends strongly on the choice of prompting strategy---e.g., the prompt text, description of robot's environment, in-context examples, etc.---and \gls{LLM} model, since choosing different prompts or \glspl{LLM} can result in varied performance when deployed, particularly when the deployment-time environments differ from those that were considered when designing such prompts.
As such, selecting only a single prompting strategy or \gls{LLM} in advance will not always elicit the best deployment-time performance.
Instead, the robot should be able to choose from more than one prompting strategies or \glspl{LLM} and evaluate each of those to pick the best ones during deployment.
However, the process of deploying and repeatedly trying out prompting strategies or \glspl{LLM} until a clear winner emerges can be problematically time consuming in general, requiring many trials to choose between them.
Recent work in the space of point-goal navigation~\citep{paudel2023selection} presents \emph{offline alt-policy replay}, in which model-based counterfactual reasoning can be used to afford choosing the best of a family of learning-informed navigation policies, a strategy we seek to leverage for prompt and \gls{LLM} selection.

To achieve effective object search performance, we therefore require a model-based approach that both informs and is informed by an \gls{LLM}: with which we can plan using the commonsense world knowledge of \glspl{LLM} and also introspect during deployment so as to quickly allow the system to select the best performing prompting strategy or \gls{LLM}.
It is a key insight of this work that a model-based planning framework for \gls{LLM}-informed object search in partially-known environments can be built upon the same high-level action abstraction used in similar approaches designed for learning-informed planning under uncertainty~\citep{stein2018learning}.
In addition, the same high-level action abstraction also affords \emph{offline replay} \citep{paudel2023selection}, and so can facilitate deployment-time evaluation of prompts and \glspl{LLM} for object search tasks---the outcomes of which can then be used to quickly select the best performing prompts and \glspl{LLM} for object search tasks.

In this work, we present an \gls{LLM}-informed model-based planning framework for object search in partially-known environments, and an accompanying approach for deployment-time selection of the best prompts and \glspl{LLM} for such \gls{LLM}-guided object search tasks (\cref{fig:overview}).
Our model-based planning framework leverages a frequently-used high-level action abstraction where the robot's actions correspond to revealing unsearched/unopened containers or furniture to look for a target object and leverages an \gls{LLM} to make predictions about statistics of uncertainty---namely, likelihood of finding an object of interest in a location---to \emph{inform}, rather than \emph{replace}, model-based object search.
Further leveraging this abstraction, we enable \emph{fast deployment-time selection} of prompts and \glspl{LLM}, a capability unique in this domain, by leveraging the \emph{offline replay} approach of \citet{paudel2023selection}.
Our contributions are as follows:
\begin{itemize}
    \item We identify a high-level action abstraction as a key enabler of both \emph{model-based planning} with \glspl{LLM} and \emph{introspection}, with which such \gls{LLM}-based systems can self-evaluate deployment-time behaviors.
    \item We present a novel approach for \gls{LLM}-informed model-based high-level planning for object search in partially-known environments that integrates predictions about uncertainty from \glspl{LLM} and known traversal costs from the occupancy map.
    \item Leveraging the action abstraction upon which planning relies, we demonstrate fast bandit-like selection of best prompts and \glspl{LLM} from a family of candidate prompts and \glspl{LLM} used for guiding the robot behavior in object search tasks.
\end{itemize}

Experiments in simulated \gls{ProcTHOR} environments demonstrate that our \gls{LLM}-informed model-based planning framework for object search outperforms \gls{LLM}-based baselines that directly ask an \gls{LLM} what action to pick and optimistic baselines, resulting up to 11.8\% and 39.2\% improvements respectively.
In addition, our prompt selection approach enables quick selection of best prompts and \glspl{LLM} from a family of prompts and \glspl{LLM}, resulting in 6.5\% lower average cost and 33.8\% lower average cumulative regret over baseline \gls{UCB} bandit selection.
Real-robot experiments with a LoCoBot robot in an apartment building show improvements for both our \gls{LLM}-informed model-based planning approach and our prompt selection approach.

\section{Related Work}
\noindent
\textbf{\Glsxtrshortpl{LLM} and \Glsxtrshortpl{VLM} for Object Search}\quad
Many recent works have explored the use of \glspl{LLM} and \glspl{VLM} and for object search tasks~\citep{dorbala2023can,zhou2023esc,yu2023l3mvn,arjun2024cognitive,ge2024commonsense,rajvanshi2024saynav,hossain2024enhancing}.
These works use \glspl{LLM} or \glspl{VLM} for their commonsense world knowledge to decide where to search~\citep{zhou2023esc,dorbala2023can,yu2023l3mvn}.
However, as they do not use a planning framework, they are often fairly myopic in their search strategies.
Our work focuses on leveraging the commonsense knowledge from \glspl{LLM} to \emph{inform} rather than \emph{replace} planning to enable reasoning about long-horizon impacts of robot's search actions: a capability important to achieve good performance for object search in partially-known environments.

\noindent
\textbf{\Glsxtrshortpl{LLM} and Planning} \quad
\Glspl{LLM} have been widely used as planners~\citep{irpan2022can,rajvanshi2024saynav,silver2022pddl,song2023llm,wu2024selp} or to augment planning~\citep{guan2023leveraging,hazra2024saycanpay,zhao2024large,liu2023llmp,nayak2024long,zhang2025lamma,ling2025elhplan}.
\Glspl{LLM} have been recently used to directly solve task planning problems in \gls{PDDL}, but their abilities to generate feasible or correct solutions are brittle~\citep{valmeekam2022large,valmeekam2023planning}.
Some recent approaches therefore aim to integrate classical planning methods with \glspl{LLM}~\citep{guan2023leveraging,liu2023llmp,hazra2024saycanpay,nayak2024long,zhang2025lamma,ling2025elhplan}.
Our work is specifically focused on extracting statistics of high-level exploratory actions from \glspl{LLM} to inform a model-based planning framework.

\noindent
\textbf{Prompt Selection}\quad
Prompt selection, which falls under a broader area of prompt engineering~\citep{sahoo2024systematic,liu2023pre}, deals with selecting the prompts that achieve the best \gls{LLM} performance on downstream tasks~\citep{yang2023improving}.
While there are approaches that aim to select the best prompts from predesigned templates~\citep{liao2022zero,liu2023pre,sorensen2022information,yang2023improving,paudel2025deploymenttime}, these approaches focus on selecting prompts that gets the best responses from \glspl{LLM} on various benchmarks and hence are not suitable for deployment-time selection of prompts in \gls{LLM}-informed object search tasks, the focus of this work.

\section{Problem Formulation}
\noindent
\textbf{Object Search in Partially-Known Environments}\quad
Our robot is tasked to find a target object $g$ in a household environment in minimum expected cost, measured in terms of distance traveled.
The environment consists of rooms, containers and objects.
\emph{Containers} are entities in the environment that can contain other objects: bed, dresser, countertop, etc.
The containers are located in different rooms in the household environment.
The belief state $b_t = \{m_t, q_t\}$ consists of the map $m_t$---with \emph{a priori} known locations of rooms and containers but what objects exist in the containers are not known---and the robot pose $q_t$, both at time $t$.
The robot must navigate to containers and search them to look for the target object.
Unexplored containers form the robot's action space $\mathcal{A}$ and the robot's policy $\pi$ maps the belief state $b_t$ to a container search action $a_t \in \mathcal{A}(b_t)$.
Our search policies are informed by \glspl{LLM} and so depend upon the choice of \glspl{LLM} and prompts used to query the \glspl{LLM}.

We presume that the robot has access to a low-level navigation planner and controller that can be used to move about and interact with the environment.
As such, the aim of our planner is to determine the sequence of container search actions that minimizes the expected cost of finding the target object.
The performance of the robot during deployment is measured as the average distance traveled by the robot to find the target object across a sequence of trials, where each trial is held in a distinct map to find an object sampled uniformly at random from the environment.

\noindent
\textbf{Prompt Selection} \quad
We consider that the robot's policy has access to multiple prompt templates and \glspl{LLM} each represented as $\theta = (\mathcal{P}, \mathcal{N})$ where $\mathcal{P}$ denotes prompt template and $\mathcal{N}$ denotes \gls{LLM}.
As such, the robot has access to a family of search policies $\Pi = \{\pi_{\theta_1}, \pi_{\theta_2}, \cdots, \pi_{\theta_N}\}$ each with a unique prompt-\gls{LLM} pair.
The objective of prompt selection is to pick the policy with a prompt-\gls{LLM} pair $\theta$ whose corresponding search actions result in minimum expected cost of finding target objects during deployment over multiple trials in distinct partially-known environments:
\begin{equation} \label{eq:prompt_selection}
    \pi^*_\theta = \argmin_{\pi_\theta \in \Pi} \mathbb{E}[C(\pi_\theta)]
\end{equation}
where $\mathbb{E}[C(\pi_\theta)]$ is the expected cost incurred by the robot upon using policy $\pi_\theta$ with a prompt-\gls{LLM} pair $\theta$ during deployment.
This problem can be formulated as a multi-armed bandit problem~\citep{sutton2018reinforcement}, solved via black-box selection algorithms like \gls{UCB}~\citep{ucb_lai1985asymptotically} using \cref{eq:ucb_bandit}:
\begin{equation} \label{eq:ucb_bandit}
    \pi^{\text{\tiny(k+1)}}_\theta = \argmin_{\pi_\theta \in \Pi} \Bigg[\,\, \bar{C}_k(\pi_\theta) - c\sqrt{\frac{\ln{k}}{n_k(\pi_\theta)}}\,\, \Bigg]
\end{equation}
where $\bar{C}_k(\pi_\theta)$ is the average cost over trials 1-through-$k$ in which policy $\pi_\theta$ with prompt-\gls{LLM} pair $\theta$ was selected, $n_k(\pi_\theta)$ is the number of times policy $\pi_\theta$ was selected until trial $k$, and $c>0$ is a parameter controlling the balance between exploration and exploitation.
However, such approaches can be slow to converge, requiring the robot to go through multiple trials of poor performance before the best policies can be identified.
White-box approaches can accelerate selection~\citep{paudel2023selection,paudel2024multistrategy}, but rely on planning abstractions that support counterfactual reasoning about robot behavior.
It is our insight that \gls{LLM}-informed planning strategies can be made compatible with such approaches and so can afford prompt and \gls{LLM} selection in this setting.

\section{\Gls{LLM}-informed Model-based Planning for Object Search} \label{sec:llm_lsp}
Here, we introduce our approach for \gls{LLM}-informed model-based planning for object search before we discuss prompt selection in \cref{sec:prompt-selection}.

We want an approach that can leverage the commonsense world knowledge of \glspl{LLM} to \emph{inform} model-based planning rather than using \glspl{LLM} to replace planning altogether.
To achieve this, we introduce an approach for \gls{LLM}-informed model-based object search, in which we seek to perform model-based planning wherein planning is augmented by the predictions generated by an \gls{LLM} about the object locations.

Our approach takes inspiration from the \gls{LSP} approach of \citet{stein2018learning}.
Their approach, designed around the aim of effective long-horizon point-goal navigation, is centered around using learning to estimate statistics associated with temporally extended actions for exploration; a learned model, trained in environments similar to those the robot sees when deployed, estimates the goodness of each such exploratory action and the likelihood that exploring the space that the action corresponds to will reach the unseen goal.

Our \gls{LLM}-informed model-based object search framework adopts a similar planning abstraction.
In this framework, each high-level action corresponds to searching the \emph{containers}, which are entities in the environment that contain other objects: bed, dresser, countertop, etc.
A search policy $\pi$ specifies the sequence of search actions the robot intends to take to find the target object.
Each such search action $a_t \in \mathcal{A}(b_t)$ has an immediate cost of first traveling to the container---corresponding to a distance $D(b_t, a_t)$ computed via A* from the occupancy grid---and then searching the container for the target object, which has a (known) search cost $R_\text{search}(b_t, a_t)$.
With a probability $P_S$, the container contains the target object and so the corresponding search action successfully finds the object.
Otherwise, with probability $1-P_S$, searching continues in other containers after picking another container search action (\cref{fig:overview}(a)).
The expected cost of a search action $a_t$ under policy $\pi$ is computed using a Bellman equation:
\begin{multline} \label{eq:bellman}
Q_\pi(b_t, a_t \in \mathcal{A}(b_t)) = D(b_t, a_t) + R_{\text{search}}(b_t, a_t)\\ + (1 - P_S(a_t)) Q_\pi(b'_t, \pi(b'_t))
\end{multline}
The robot's policy $\pi(b_t) \equiv \argmin_a Q_\pi(b_t, a \in \mathcal{A}(b_t))$ can be used to compute a search plan: the sequence of actions that minimizes the expected cost via \cref{eq:bellman} to find the target object.
To speed up planning, we limit the action space using heuristics to choose up to eight containers with high likelihoods $P_S$ and low travel costs $D$ and select among them the container with lowest expected cost, incorporating additional containers as the robot moves and searches.
It should be noted that our planning strategy is complete since, explored containers are removed from candidate action set and so in the worst case, the robot explores all available containers to find the target object if it exists.
Using an \gls{LLM} as the knowledge repository of where common objects of interest might be located in the given environment, we prompt the \gls{LLM} provide an estimate of the marginal probability $P_S$
and use it to compute the expected cost via \cref{eq:bellman}.
Such augmentation of planning with predictions from \glspl{LLM} enables effective reasoning without explicitly relying on \glspl{LLM} for multi-step reasoning thereby enabling improved performance.

\section{Prompt Selection for \Glsxtrshort{LLM}-informed Object Search} \label{sec:prompt-selection}
\subsection{Overview of Prompt Selection}
When using an \gls{LLM} to inform planning for object search, prompts used to query \glspl{LLM} for object likelihood predictions would ideally result in effective performance.
However, effectiveness of a plan in the context of object search in partially-known environments can only be realized after the robot executes them in the environments.
Thus, a poor prompting strategy may only be identified as such after the robot deploys and relies upon that \gls{LLM} and prompt combination---a costly strategy of trial-and-error for robot navigation tasks.
Instead, if we could identify poor prompts while limiting the need to deploy them during deployment, we could rule them out quickly and prioritize selection of the best prompts to enable improved robot performance.

It is a key insight of this work that \emph{offline replay} approach by \citet{paudel2023selection} can be used to select between prompts without the robot having to deploy the plans informed by \glspl{LLM} using such family of prompts.
While used by \citet{paudel2023selection} in the context of point-goal navigation in partially-mapped environments to replay the behavior of alternative policies without having to deploy them, we adapt offline replay to determine what the robot would have done if it had instead used a different prompt or \gls{LLM} to guide its behavior.
Costs from offline replay (\cref{fig:offline-replay-overview}) of alternative prompts and \glspl{LLM}, $\bar{C}^{\text{\tiny{rep}}}_k$ (averaged over trials 1-through-$k$) can then be used in \gls{UCB} bandit-like selection strategy (\cref{fig:overview}(b)) similar to that of \citet{paudel2023selection} to pick the policy $\pi_\theta$ with prompt-\gls{LLM} pair $\theta$ for trial $k+1$ as:
\begin{equation}\label{eq:offline_replay_selection}
    \pi^{\text{\tiny(k+1)}}_\theta = \argmin_{\pi_\theta \in \Pi} \Bigg[\max \Bigg({\bar{C}^{\text{\tiny{rep}}}_k(\pi_\theta),  \bar{C}_k(\pi_\theta) - c\sqrt{\frac{\ln{k}}{n_k(\pi_\theta)}} }\,\Bigg)\Bigg]
\end{equation}

\subsection{Object Search during a Trial} \label{sec:object_search_trial}
In each trial $k$ when the robot is deployed in a partially-known environment to look for a target object $g$, it uses \cref{eq:offline_replay_selection} to choose one of the policies $\pi^{\text{\tiny{(k)}}}_\theta \in \Pi$ defined by its prompt-\gls{LLM} pair $\theta=(\mathcal{P}, \mathcal{N})$, uses the prompt $\mathcal{P}$ to query the \gls{LLM} $\mathcal{N}$ for object likelihood $P_S$ and computes the next action $a_t$ corresponding to searching one of the unexplored containers using \cref{eq:bellman}.
The robot searches the container corresponding to action $a_t$ to look for the target object $g$, repeating planning and search each time the target object is not found.
The total distance traveled by the robot to find the object is the cost $C_k(\pi_\theta)$ for trial $k$.
The robot stores the information $\mathcal{Z}_k$ about the contents of all containers it explored and the known map of the environment to be used later for offline replay (\cref{sec:offline-replay-prompt}).

\subsection{Prompt Selection with Offline Replay of Alternative Prompts and \Glspl{LLM}} \label{sec:offline-replay-prompt}

\begin{figure}[t]
    \vspace{0.5em}
    \centering
    \includegraphics[width=0.95\linewidth]{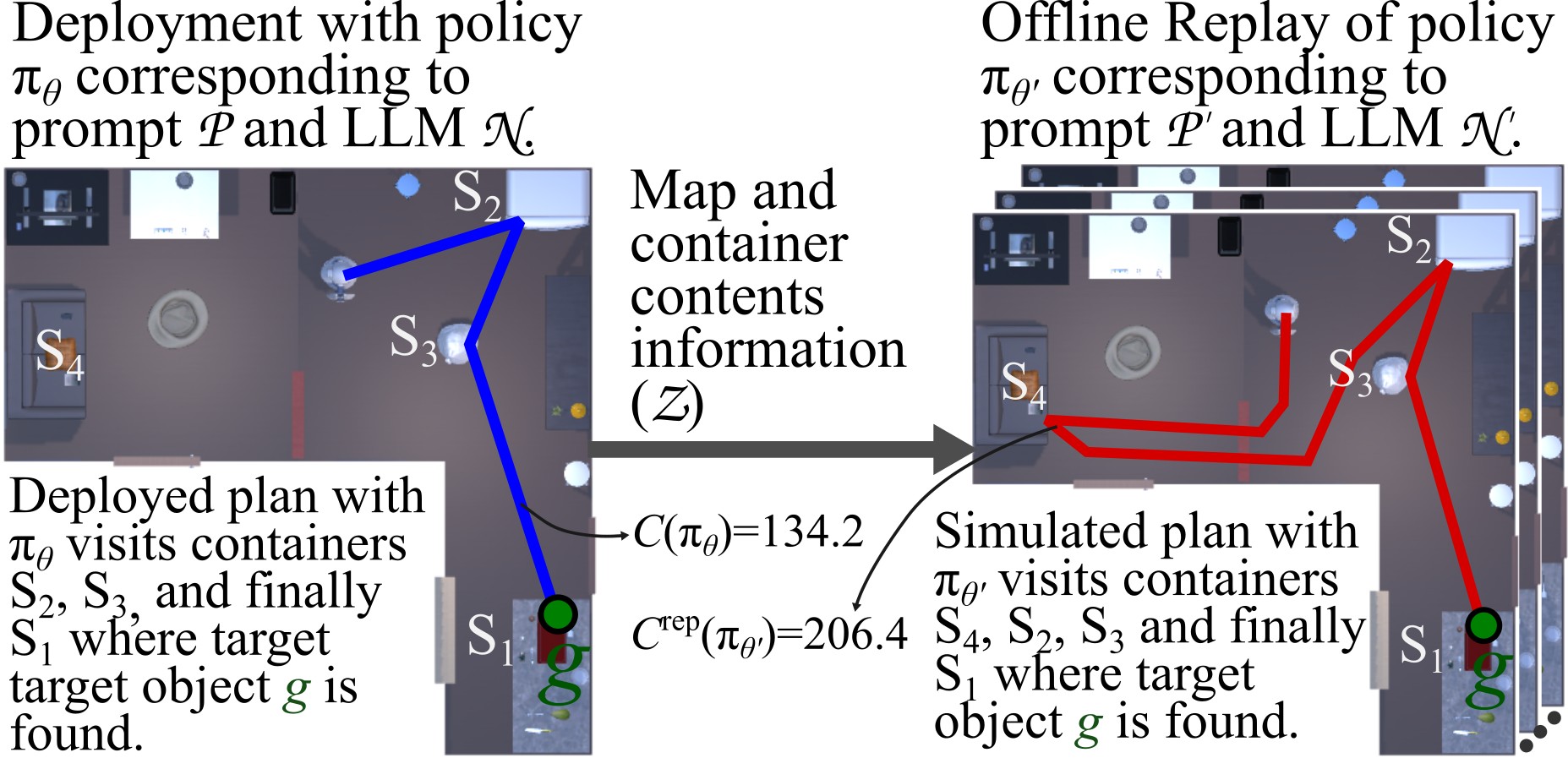}
    \vspace{-0.7em}
    \caption{Overview of offline replay of alternative prompts and \glspl{LLM}}
    \vspace{-1.8em}
    \label{fig:offline-replay-overview}
\end{figure}

After a trial $k$ is complete, our robot uses policy $\pi_{\theta'}$ with alternative prompt-\gls{LLM} pair $\theta'$ and computes an alternative search plan.
However, deploying such alternative plans is expensive in general.
Instead, we use the hindsight information $\mathcal{Z}_k$ about the location of the target object found in trial $k$ and the existing map to \emph{replay} what the robot would have done if it had deployed a plan corresponding to an alternative prompt-\gls{LLM} pair $\theta'= (\mathcal{P}', \mathcal{N}')$ (\cref{fig:offline-replay-overview}).
Since we now know in advance which container contained the target object $g$ based on the information $\mathcal{Z}_k$ and pessimistically
assume that all other containers would not have contained the target object, we can compute the cost of following a separate policy: the length of the trajectory the robot would have taken by following the alternate search policy to find that target object.
The offline replay of a policy resembles actual deployment: replanning is done until the replayed policy suggests searching the container where the target object was found during deployment, accumulating traversal costs along the way.
The average cost over trials of the offline-replayed plans, $\bar{C}^{\text{\tiny{rep}}}_k(\pi_{\theta'})$, for an alternative prompt-\gls{LLM} pair $\theta'$ and average costs over trials of the chosen prompt-\gls{LLM} pairs $\theta$ in trial $k$, $\bar{C}_k(\pi_\theta)$ are together used to pick the prompt and \gls{LLM} in subsequent trials $k+1$ using \cref{eq:offline_replay_selection}.
While the replay costs based on pessimistic assumptions may be somewhat biased, this has been empirically shown to more tightly constrain bandit-like selection compared to other relaxed assumptions in prior work of \citet{paudel2023selection} that inspired our offline replay approach.

\section{Simulation Experiments and Results} \label{sec:simulation_experiments}
\subsection{Experiment Design}

\begin{figure*}[t]
    \centering
    \vspace{0.5em}
    \includegraphics[width=\textwidth]{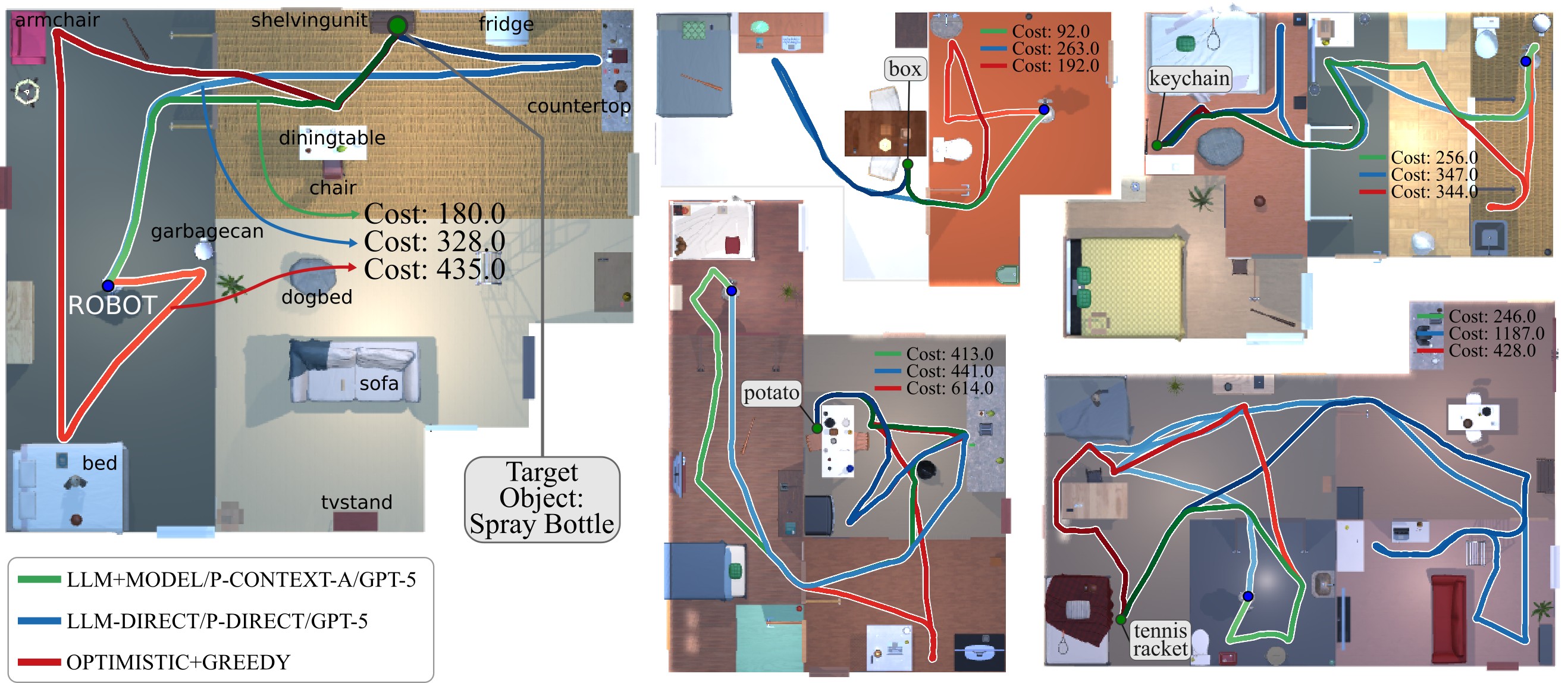}
    \caption{\textbf{Sample robot trajectories:} Using \gls{GPT}-5 as \gls{LLM}, our \ourplanner{} policy finds the target object with lower costs compared to \fulllm{} and \optimistic{} policies.}
    \label{fig:results-trajectories}
    \vspace{-2em}
\end{figure*}

We perform simulation experiments for object search in 150 distinct household environments based on the \gls{ProcTHOR}~\citep{deitke2022procthor} dataset, which consists of procedurally generated homes.
Our robot has access to the underlying occupancy grid of the environment and what containers exist in what rooms, yet the contents of the containers are not known to the robot. Containers include any entities that could contain an object and so include more traditional containers, such as cabinets or boxes, but also surfaces, including tables, countertops, and bookshelves.
The robot must travel to the container locations and search the containers to find the object of interest.
For the purposes of our experiments, we do not directly simulate the robot executing manipulation skills (for example, to open a fridge or a cabinet), and we treat objects as being instantaneously revealed upon reaching a container and so assign search cost $R_\text{search}=0$.

\noindent
\textbf{Policies}\quad
We perform object search experiments with our \ourplanner{} model-based planner discussed in \cref{sec:llm_lsp} and also design two baseline policies, \fulllm{} and \optimistic{} discussed below:
\begin{description}[leftmargin=15pt]
    \item[\ourplanner{}]
        This is our \gls{LLM}-informed model-based planner that uses an \gls{LLM} to obtain object likelihood probabilities and then uses \cref{eq:bellman} to select the best container search action as discussed in \cref{sec:llm_lsp}.
    \item[\fulllm{}]
        This \gls{LLM}-informed baseline policy directly prompts the \gls{LLM} to respond with the container the robot should search next, instead of asking for probabilities as we do with our \ourplanner{} planner.
        As such, \fulllm{} policy does not use a planning framework to compute actions and instead directly executes actions picked by the \gls{LLM} from a list of all available container search actions.
    \item[\optimistic{}]
        This non-\gls{LLM} baseline optimistically assumes that all containers could contain the target object and greedily searches the nearest container, replanning until the target object is found.
\end{description}

\noindent
\textbf{\Gls{LLM} Variants} \quad
We experiment with two \glspl{LLM}, \gls{GPT}-5 Mini and Gemini 2.5 Flash, for object search tasks in \gls{ProcTHOR} household environments.

\noindent
\textbf{Prompt Design}\quad
We construct multiple prompts to query the \glspl{LLM} for guiding the robot behavior.
The prompt design for \ourplanner{} policy and \fulllm{} policy are slightly different since for \ourplanner{} we want the \gls{LLM} to generate probability values for each container, while for the \fulllm{} policy, the \gls{LLM} should directly output which container the agent should search next.
While such prompts might be constructed with variations in language, context and the role that \gls{LLM} should play in the interaction, each prompt includes a question asking the \gls{LLM} to respond with either a probability value (for \ourplanner{} policy) or the name of the container to search (for \fulllm{} policy).
We design three prompt templates for \ourplanner{} policy: \prompta{}, \promptb{} and \promptc{}, and one prompt template for \fulllm{} baseline policy: \promptd{}.
Examples of each are included in Appendix~\ref{app:prompts}.
\begin{description}[leftmargin=15pt]
    \item[\prompta{}, \promptb{}:]
        These prompts are designed around four main elements: (i) a description of the setting and the role that the \gls{LLM} will serve, (ii) a description of the house including a list of the rooms present and the containers they contain, (iii) an example for reference, and (iv) the query asking for the probability of finding the object of interest in a container within a particular room.
        While the semantic meaning of these prompts are similar (see Appendix~\ref{app:prompts}), each of these differ in terms of the language is used in the prompt text.
    \item[\promptc{}:]
        This prompt doesn't include any of the aforementioned contexts about the \gls{LLM}'s role, environment description and reference example, and only includes the query asking for the probability of finding the target object in a container within a particular room.
    \item[\promptd{}:]
        This prompt for \fulllm{} policy is designed around five main elements: (i) a description of the setting and the role that the \gls{LLM} will serve (ii) an example interaction for reference (iii) a description of the house including a list of the rooms present and the distances between them (iv) list of available containers that the robot can explore, and (v) the query asking which container the robot should explore to find the target object quickly.
        It should be noted that we include the distances in the prompt because we expect the \gls{LLM} to behave like a planner and so provide all necessary information needed to plan effectively.
\end{description}

\noindent
\textbf{Prompt Selection} \quad
We compare the  prompt/\gls{LLM} selection approach that leverages a high-level action abstraction to facilitate offline replay as discussed in Sec.~\ref{sec:prompt-selection}, referred to as Replay Selection, against a baseline black-box \gls{UCB} bandit selection approach.
Since our selection approach is enabled by our high-level action abstraction, our other strategies that do not explicitly use planning or query \glspl{LLM}---yet use the same underlying abstraction---are also amenable to offline replay, and hence can be included as a part of candidate strategies for bandit-like selection.
As such, for our experiments, selection seeks to choose between all nine strategies for object search from \cref{tab:results-single-prompt}.
Each deployment lasts for 100 trials, each in a distinct \gls{ProcTHOR} map, over which selection proceeds.
While the \gls{UCB} Selection uses only the deployment cost of a strategy to pick the policy-prompt-\gls{LLM} combination for subsequent trials using \cref{eq:ucb_bandit}, Replay Selection additionally uses the offline replay costs of all other policy-prompt-\gls{LLM} combination to pick the strategy for next trial via \cref{eq:offline_replay_selection}.

\subsection{Policy Performance Results}
\begin{table}[t]
    \centering
    \footnotesize
    \vspace{0.5em}
    \caption{Navigation Costs for Object Search Tasks}
    \label{tab:results-single-prompt}
    \begin{tabular}{lc}
    \toprule
    \textbf{Policy / Prompt / \Gls{LLM}}     &  \textbf{Avg. Nav. Cost} \\
    \midrule
    \ourplanner{} (ours) / \prompta{} / \gls{GPT}-5     & \textbf{242.38} \\
    \ourplanner{} (ours) / \promptb{} / \gls{GPT}-5     & 252.14 \\
    \ourplanner{} (ours) / \promptc{} / \gls{GPT}-5     & 263.94 \\
    \fulllm{} (baseline) / \promptd{} / \gls{GPT}-5     & 274.94 \\
    \midrule
    \ourplanner{} (ours) / \prompta{} / Gemini     & 224.70 \\
    \ourplanner{} (ours) / \promptb{} / Gemini     & \textbf{215.59} \\
    \ourplanner{} (ours) / \promptc{} / Gemini     & 266.20 \\
    \fulllm{} (baseline) / \promptd{} / Gemini     & 233.41 \\
    \midrule
    \optimistic{} (baseline) / -- / --     & 354.34 \\
    \bottomrule
    \end{tabular}
    \vspace{-1.9em}
\end{table}

\begin{figure}[t]
    \centering
    \footnotesize
    \vspace{0.5em}
    \begin{tabular}{@{\hspace{1pt}}l@{\hspace{4pt}}l@{\hspace{5pt}}c@{\hspace{5pt}}c@{\hspace{5pt}}c@{\hspace{1pt}}}
        \toprule
        \multirow{2}{*}{\textbf{\begin{tabular}[c]{@{}l@{}}Metric\end{tabular}}} & \multirow{2}{*}{\textbf{\begin{tabular}[c]{@{}l@{}}Selection\\ Approach\end{tabular}}} & \multicolumn{3}{c}{\textbf{Num of Trials ($k$)}} \\ \cmidrule(l){3-5}
                                                                                          &                            & $k=20$           & $k=50$          & $k=100$         \\
        \midrule
        \multirow{2}{*}{\begin{tabular}[c]{@{}l@{}}Avg.\\ Cost\end{tabular}}              & UCB Selection              & 258.72\textcolor{taborange}{$\blacktriangle$}         & 254.63\textcolor{taborange}{$\blacklozenge$}        & 247.25\textcolor{taborange}{$\blacksquare$}        \\
                                                                                          & Replay Selection (ours)    & \textbf{248.12}\textcolor{tabblue}{$\blacktriangle$}         & \textbf{240.40}\textcolor{tabblue}{$\blacklozenge$}        & \textbf{231.29}\textcolor{tabblue}{$\blacksquare$}        \\
        \midrule
        \multirow{2}{*}{\begin{tabular}[c]{@{}l@{}}Cumul.\\ Regret\end{tabular}}          & UCB Selection              & 884.7\textcolor{taborange}{$\pmb{\vartriangle}$}          & 2102.8\textcolor{taborange}{$\pmb{\lozenge}$}        & 3800.2\textcolor{taborange}{$\pmb{\square}$}        \\
                                                                                          & Replay Selection (ours)    & \textbf{751.0}\textcolor{tabblue}{$\pmb{\vartriangle}$}          & \textbf{1579.8}\textcolor{tabblue}{$\pmb{\lozenge}$}        & \textbf{2516.5}\textcolor{tabblue}{$\pmb{\square}$}        \\
        \bottomrule
        \vspace{-6pt}
    \end{tabular}
    \includegraphics[width=0.8\linewidth]{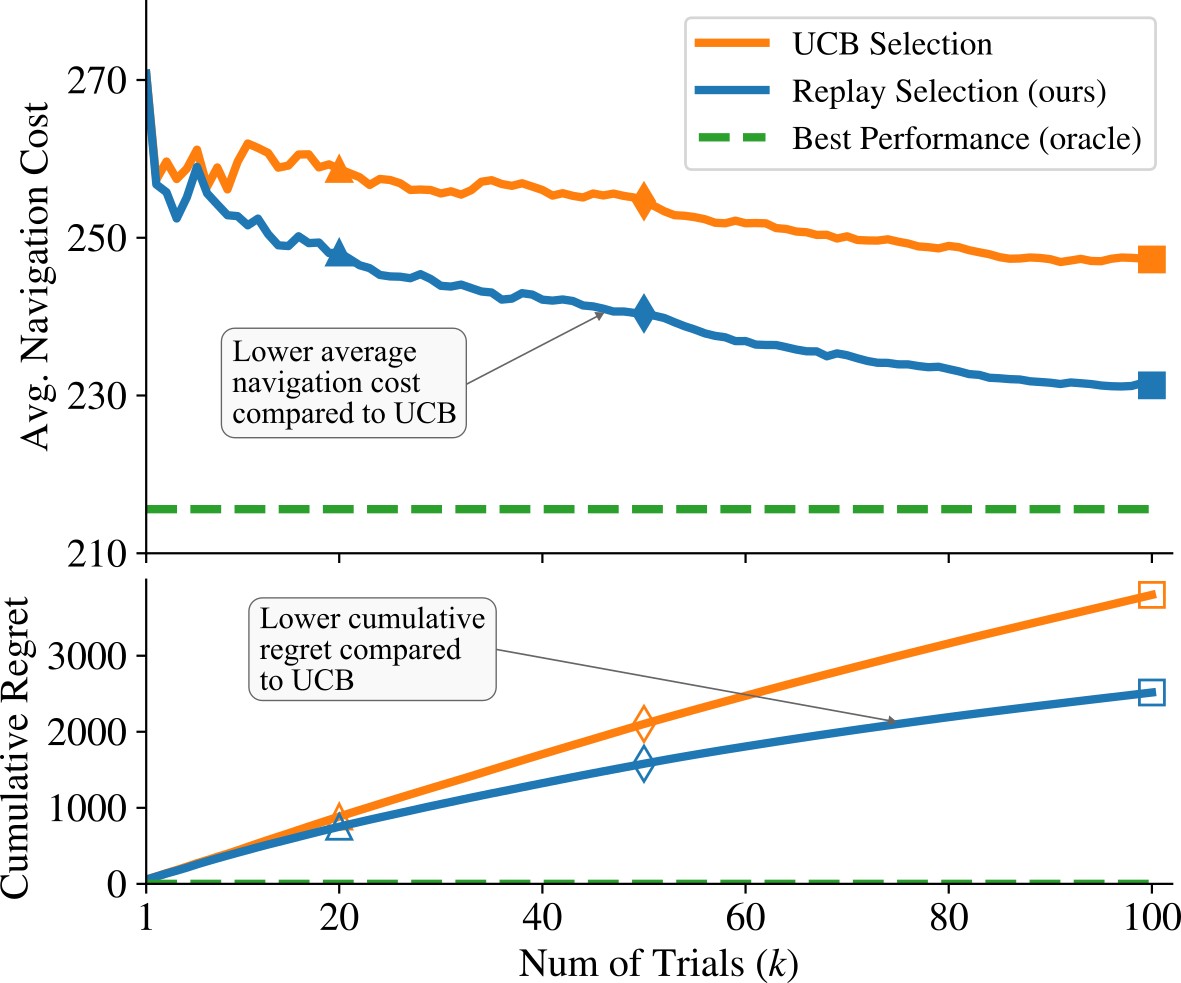}
    \vspace{-0.7em}
    \caption{\textbf{Prompt-\Gls{LLM} Selection Results}: Leveraging offline replay for prompt/model selection allows faster selection of the best prompting strategy compared to the \gls{UCB}.}
    \vspace{-2.3em}
    \label{fig:results-prompt-selection}
\end{figure}

We deploy different combinations of policies, prompts and \glspl{LLM} (the state of the art \gls{GPT}-5 and Gemini 2.5) each in 150 distinct maps from the \gls{ProcTHOR} household environments, matching our discussion above.
The average navigation costs for each such combinations are shown in \cref{tab:results-single-prompt}.
We also include example robot trajectories for three of the representative policies in \cref{fig:results-trajectories}.

\noindent
\textbf{Results with the \gls{GPT}-5 \gls{LLM}}\quad
When using \gls{GPT}-5 as the \gls{LLM} to estimate object discovery likelihoods and so guide robot behavior, our model-based \ourplanner{} planners outperform the fully-\gls{LLM}-based \fulllm{} policy, with our improvements between 4.0--11.8\%, showing that our \gls{LLM}-informed model-based planning approach enables considerable improvements in object search performance in partially-known environments.
Additionally, both the \fulllm{} \gls{LLM}-based baseline and our \ourplanner{} policies outperform the uninformed \optimistic{} baseline policy as they can both leverage the commonsense reasoning of the \gls{LLM} to search more effectively.
Our \ourplanner{} planners demonstrate improvements between 25.5--31.6\% over the \optimistic{} policy.

\noindent
\textbf{Results with the Gemini \gls{LLM}}\quad
Using Gemini as the \gls{LLM} for guiding the robot behavior, we outperform the \fulllm{} policy with all our model-based \ourplanner{} planners except for the planner relying on the \promptc{} prompt, which includes no context about the surrounding environment; the remaining \ourplanner{} policies improve upon \fulllm{} baseline policy, achieving up to a 7.6\% improvement.
Additionally, our \ourplanner{} planners also outperform the uninformed \optimistic{} policy with improvements between 24.9--39.2\%.

\noindent
\textbf{Discussion} \quad
We observe that our \gls{LLM}-informed model-based planning approach outperforms those that fully relies on \gls{LLM} to pick robot's search actions.
These results highlight the importance of using a model-based planning framework in tandem with \glspl{LLM}, rather than using \glspl{LLM} in place of planners, to benefit from the strengths of both---a key focus of this work.
Moreover, we observe that even the same prompts used with different \glspl{LLM} can result in significantly different performances: for \gls{GPT}-5, \prompta{} prompt outperforms all other strategies, and for Gemini, \promptb{} prompt outperforms all other strategies.
This result highlights that relying on a single prompt or \gls{LLM} may not always yield the best performance and as such, one must select the best prompt and \gls{LLM} combination \emph{during deployment} to maximize good performance in the environment the robot is deployed in.

\begin{table*}[th]
\centering
\footnotesize
\vspace{0.5em}
\caption{\textbf{Real-world LLM-informed Object Search Results:} Our \ourplanner{} policy outperforms baseline \fulllm{} policy.}
\label{tab:real_robot_results}
\begin{tabular}{@{}lcccccc@{}}
\toprule
\multirow{2}{*}{ \textbf{Policy / Prompt / \gls{LLM}}}                                      & \multicolumn{5}{c}{\textbf{Target Object}}                       & \multirow{2}{*}{\begin{tabular}[c]{@{}c@{}}\textbf{Average Cost}\\ \textbf{(meters)}\end{tabular}} \\ \cmidrule(lr){2-6}
                                                                                                   & \textit{blanket} & \textit{cell phone} & \textit{remote} & \textit{pillow} & \textit{wallet} &                                                                         \\ \midrule
\ourplanner{} (ours) / \promptb{} / \gls{GPT}-5 & 23.5    & \textbf{24.7}       & \textbf{7.6}            & \textbf{6.8}    & \textbf{13.2}   & 15.2                                                                    \\
\ourplanner{} (ours) / \promptc{} / \gls{GPT}-5 & \textbf{23.2}    & \textbf{24.7}       & \textbf{7.6}            & \textbf{6.8}    & \textbf{13.2}   & \textbf{15.1}                                                                    \\
\fulllm{} (baseline) / \promptd{} / \gls{GPT}-5 & 23.7    & 33.6       & 14.3           & \textbf{6.8}    & 27.9   & 21.3                                                                    \\ \bottomrule
\vspace{-3.0em}
\end{tabular}
\end{table*}

\begin{figure}[t]
    \centering
    \includegraphics[width=0.74\linewidth]{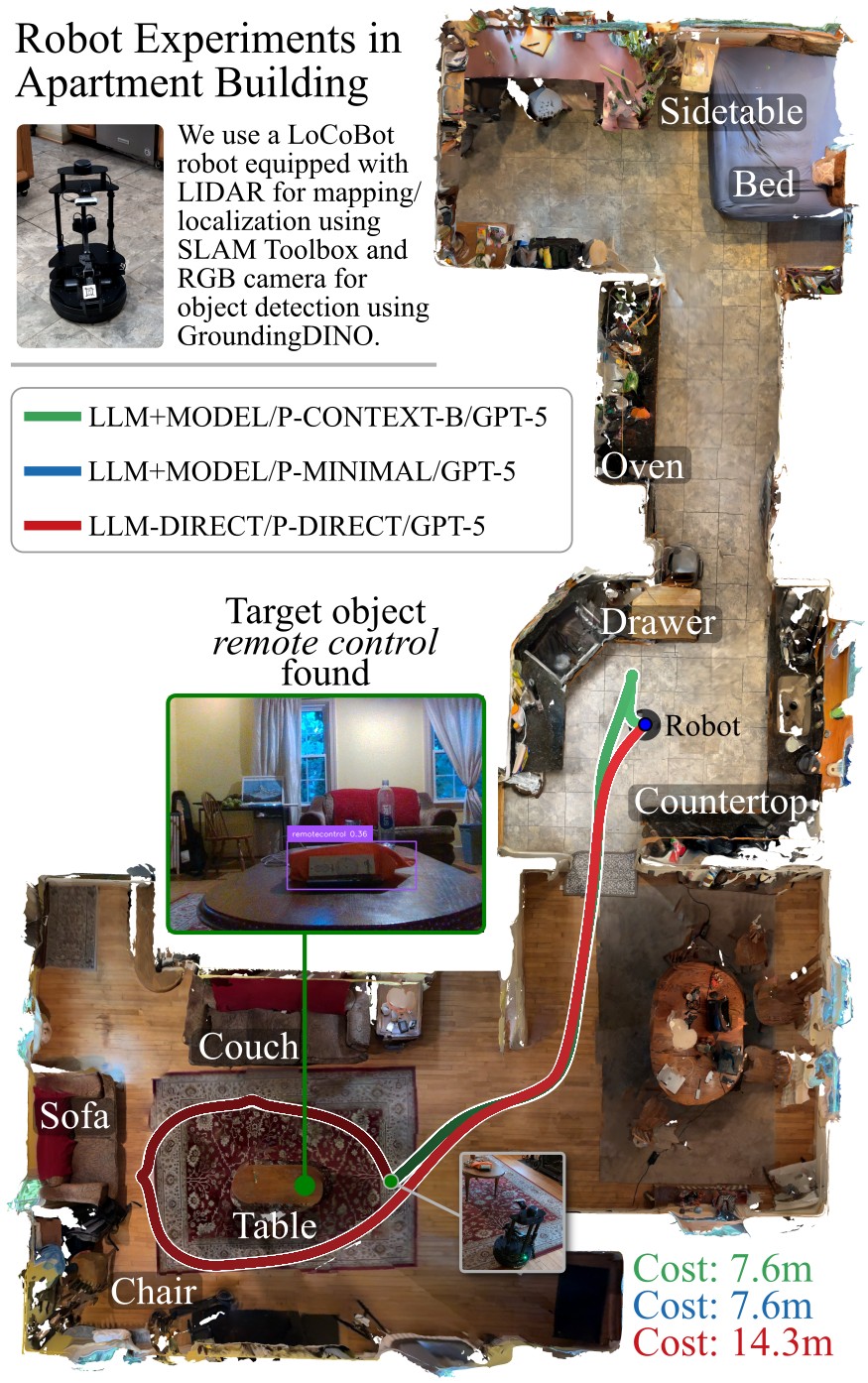}
    \vspace{-1.0em}
    \caption{\Gls{LLM}-informed Object Search Experiment in an Apartment}
    \vspace{-1.5em}
    \label{fig:real_robot_results}
\end{figure}

\begin{figure}[t]
    \centering
    \footnotesize
    \begin{tabular}{@{\hspace{3pt}}l@{\hspace{5pt}}lc@{\hspace{6pt}}c@{\hspace{6pt}}c@{\hspace{3pt}}}
        \toprule
        \multirow{2}{*}{\textbf{\begin{tabular}[c]{@{}l@{}}Metric\end{tabular}}} & \multirow{2}{*}{\textbf{\begin{tabular}[c]{@{}l@{}}Selection\\ Approach\end{tabular}}} & \multicolumn{3}{c}{\textbf{Num of Trials ($k$)}} \\ \cmidrule(l){3-5}
                                                                                          &                            & $k=2$           & $k=3$          & $k=5$         \\
        \midrule
        \multirow{2}{*}{\begin{tabular}[c]{@{}l@{}}Avg.\\ Cost\end{tabular}}              & UCB Selection              & 28.6\textcolor{taborange}{$\blacktriangle$}         & 21.6\textcolor{taborange}{$\blacklozenge$}        & 17.0\textcolor{taborange}{$\blacksquare$}        \\
                                                                                          & Replay Selection (ours)    & \textbf{24.1}\textcolor{tabblue}{$\blacktriangle$}         & \textbf{18.6}\textcolor{tabblue}{$\blacklozenge$}        & \textbf{15.2}\textcolor{tabblue}{$\blacksquare$}        \\
        \midrule
        \multirow{2}{*}{\begin{tabular}[c]{@{}l@{}}Cumul.\\ Regret\end{tabular}}          & UCB Selection              & 21.9\textcolor{taborange}{$\pmb{\vartriangle}$}          & 28.3\textcolor{taborange}{$\pmb{\lozenge}$}        & 33.0\textcolor{taborange}{$\pmb{\square}$}        \\
                                                                                          & Replay Selection (ours)    & \textbf{17.5}\textcolor{tabblue}{$\pmb{\vartriangle}$}          & \textbf{21.0}\textcolor{tabblue}{$\pmb{\lozenge}$}        & \textbf{21.6}\textcolor{tabblue}{$\pmb{\square}$}        \\
        \bottomrule
        \vspace{-2pt}
    \end{tabular}
    \includegraphics[width=0.8\linewidth]{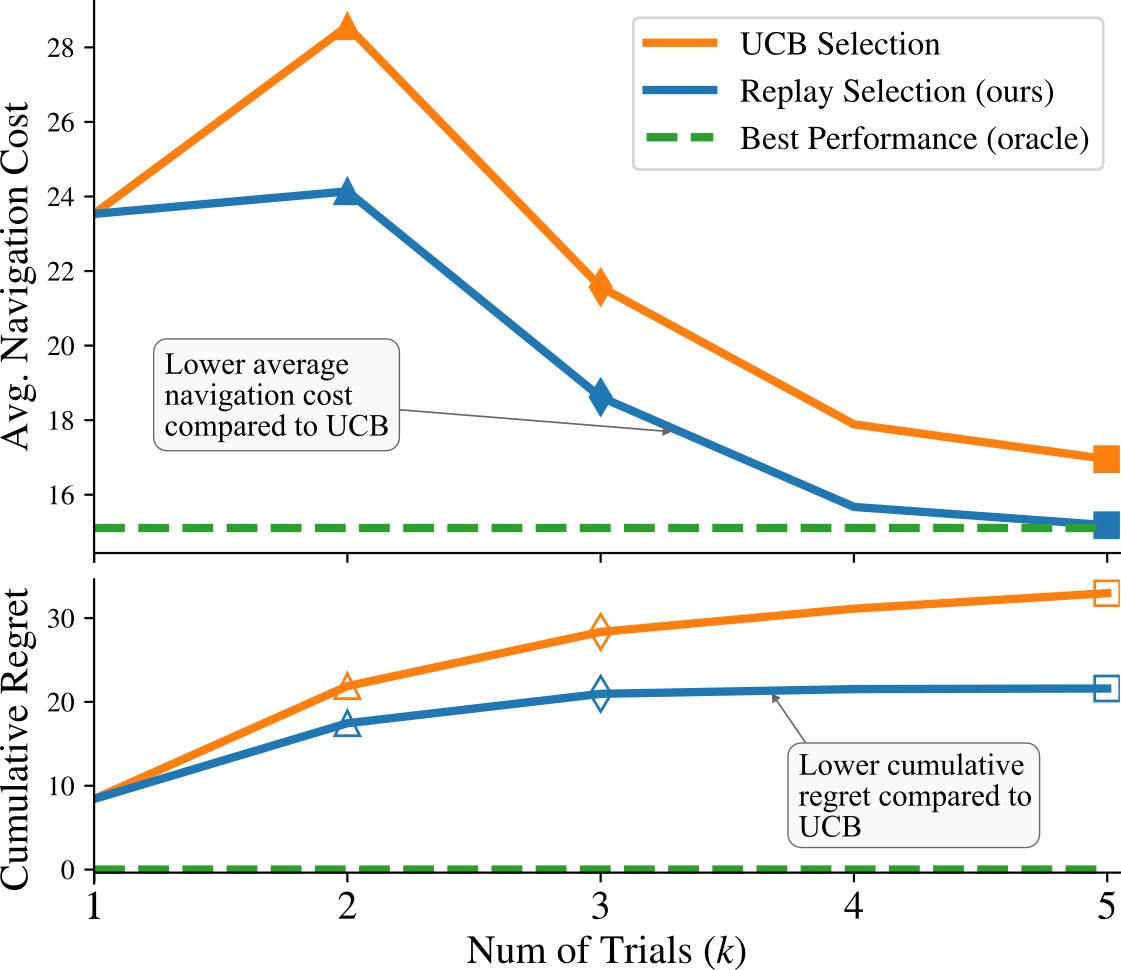}
    \vspace{-1.0em}
    \caption{\textbf{Prompt-LLM Selection in an Apartment}: Our Replay Selection allows faster selection of the best prompting strategy compared to the \gls{UCB} selection strategy, resulting in lower average cost and cumulative regret.}
    \vspace{-1.5em}
    \label{fig:prompt_selection_real_robot}
\end{figure}

\subsection{Prompt Selection Results}

As discussed in Sec.~\ref{sec:prompt-selection}, it is another key insight of our work that the action abstraction leveraged in the previous section is compatible with the \emph{offline replay selection}, which we can therefore leverage to more quickly select the best policy-prompt-\gls{LLM} combination.
To evaluate the statistical performance of selection, we generate 500 unique deployments by randomly permuting 100 trials from a set of 150 distinct maps, expecting the robot to perform selection over all nine policy-prompt-\gls{LLM} combinations shown in \cref{tab:results-single-prompt} separately for each sequence.

In \cref{fig:results-prompt-selection}, we report the \emph{average navigation cost}, which corresponds to the average of navigation costs incurred in trials 1-through-$k$, averaged over all 500 deployments.
The \emph{cumulative regret}, also shown in \cref{fig:results-prompt-selection}, tracks performance over time as the cumulative difference between the selection-based policy and a Best Performance oracle that knows in advance which strategy is best:
\ourplanner{}/\promptb{}/Gemini.
Our results demonstrate a reduction of 6.5\% in average cost at the end of 100th trial compared to a standard \gls{UCB}-bandit selection approach.
In particular, we achieve 33.8\% lower cumulative regret after 100th trial compared to \gls{UCB}-bandit selection, a number which would continue to grow with more trials.

Our results highlight the need and benefits of fast deployment-time selection of prompts and \glspl{LLM}, since without such selection, the robot risks poor performance if it uses only one prompt or \gls{LLM} preselected before deployment.

Our selection approach enables the robot to quickly pick \emph{during deployment} the prompts and \glspl{LLM} that yield better behavior and hence maximizing long-term performance---a benefit afforded by our high-level action abstraction amenable to \emph{offline replay} of \citet{paudel2023selection}.

\section{Real-Robot Demonstration} \label{sec:real_robot_results}

\noindent
\textbf{Policy Performance Results}\quad We demonstrate the effectiveness of our \gls{LLM}-informed model-based planning and prompt selection approach with a LoCoBot robot in an apartment containing a kitchen, a dining room, a living room and a bedroom.
These rooms contain a total of nine containers where the robot can search for target object as shown in \cref{fig:real_robot_results}.
We conduct five trials in which the robot starts in the kitchen and is tasked to find a distinct object in each trial.
For the purpose of demonstration, we use \gls{GPT}-5 as our \gls{LLM} and use \promptb{} and \promptc{} prompts for \ourplanner{} policy and compare with \fulllm{} policy using \promptd{} prompt.
We show one representative example in  \cref{fig:real_robot_results} and report all results in \cref{tab:real_robot_results}.
Across five trials, our \ourplanner{} policy incurs an average navigation cost of 15.1m compared to the baseline \fulllm{} policy which incurs the cost of 21.3m, thus outperforming \fulllm{} policy by 29\%.

\noindent
\textbf{Prompt Selection Results}\quad
We also deploy our prompt selection approach on the LoCoBot robot with three policy-prompt-\gls{LLM} combination used in aforementioned object search experiments for five trials in which the robot is tasked to find a distinct object.
We observe that over the course of five trials, each with distinct target object, our Replay Selection approach incurs an average cost of 15.2m  while \gls{UCB} Selection incurs an average cost of 17.0m, an improvement of 10.5\% as shown in \cref{fig:prompt_selection_real_robot}.
Similarly, our Replay Selection approach incurs a cumulative regret of 21.6m compared to UCB which incurs 33.0m, showing an improvement of 34.5\%.

\section{Conclusion}

We present a novel framework that seeks to integrate \glspl{LLM} with high-level model-based reasoning for performant object search.
Our approach first introduces an \gls{LLM}-informed model-based high-level planner for object search in partially-known environments that integrates predictions about uncertainty from \glspl{LLM} and information partially-known from the environment.
Second, by leveraging the high-level action abstraction upon which model-based planning relies, we demonstrate that the recent \emph{offline replay} approach developed for model selection for learning-informed point-goal navigation~\citep{paudel2023selection} can be made to support fast deployment-time prompt and \gls{LLM} selection, a capability unique in this domain.
In future, we hope to extend our model-based planning approach to fully unknown environments, which the robot must reveal through exploration and observation, and multi-robot settings \citep{khanal2023mrlsp,khanal2025learning,khanal2026mrtask}.
Additionally, our prompts are manually generated and remain unchanged during deployment.
In future, we hope to explore automated prompt generation and refinement strategies that integrate with our approach.
Coupled with our offline replay approach, such methods could enable reliable deployment-time refinement of prompts across diverse environments without requiring on-robot trial-and-error.

\section*{Acknowledgements}
This work was supported by the National Science Foundation (NSF) under Grant 2232733, and the U.S. Army Research Laboratory (ARL) under Grant W911NF2520011.

\bibliography{references.bib}

\appendix
\subsection{Samples of Prompts} \label{app:prompts}
\begin{figure}[t]
    \centering
    \vspace{0.6em}
    \includegraphics[width=\linewidth]{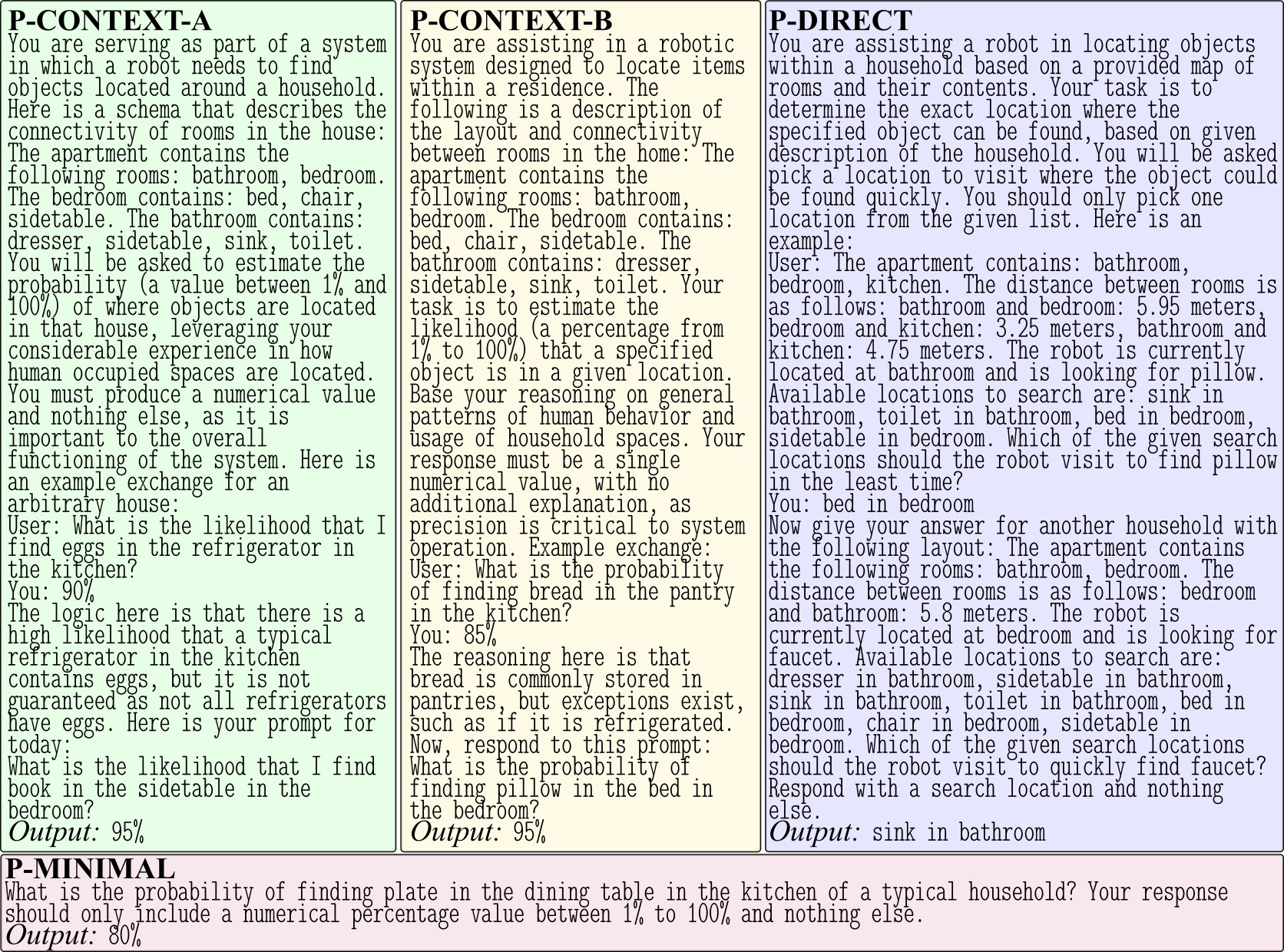}
    \vspace{-2.0em}
    \caption{Samples of prompts used in our experiments}
    \vspace{-1.5em}
    \label{fig:prompts}
\end{figure}

As discussed in \cref{sec:simulation_experiments}, we include in \cref{fig:prompts} the samples of all prompts used in our experiments.

\subsection{Details about Belief Updates} \label{app:belief_updates}
As mentioned in \cref{sec:object_search_trial}, our belief state includes the occupancy map, locations of containers, whether each container has been explored or not, and objects found in each container. Upon exploring a container, the objects found in that container are added to the robot's partial map, and the container is marked as explored. If the target object was not found in that container, planning continues with the updated set of container search actions that only includes unexplored containers, and does not include any container that has been marked as explored. This process is identical for all planning strategies considered in our work.

\subsection{Scalability in terms of Number of Containers and Prompts/\Glspl{LLM}} \label{app:scalability}
As is true with many exhaustive search-based strategies, the speed of planning scales poorly  with the number of containers or apartment size since the number of feasible plans scales factorially with the increase in number of containers. As discussed in \cref{sec:llm_lsp}, we limit the action space to speed up planning by using heuristics to choose up to eight containers with high likelihoods $P_S$ and low travel costs $D$ and select among them the container with lowest expected cost, incorporating additional containers as the robot moves and searches. Since the set of actively considered containers is updated as those containers are explored, there is no loss of generality, and this strategy works well in practice and yields effective performance as demonstrated in our experiments.

In terms of the number of prompts and \glspl{LLM}, our prompt selection approach scales linearly since the addition of a new prompt/\gls{LLM} would only add the overhead of replaying this new prompt/\gls{LLM} after a trial is complete. During each replay, the most computational overhead comes from querying the \gls{LLM}, which we additionally mitigate by caching the \gls{LLM}'s responses from deployment. As such, the replay in itself is quite inexpensive (less than a couple of seconds each), making our prompt/\gls{LLM} selection approach highly scalable to large number of prompts and \glspl{LLM}.

\subsection{Using Marginal Probabilities} \label{app:marginal_probabilities}
For each container search action, we obtain likelihood $P_S$ of finding the target object in that container by querying \glspl{LLM}. These likelihoods are not normalized across available containers because they do not need to be a valid probability mass. For each container, $P_S$ is a marginal probability that represents the likelihood of finding the target object in that container, which is treated as being independent of what other containers exist. The information about what other containers exist and their respective likelihoods are instead used by our planning approach to compute the best action thus alleviating the need to normalize these probabilities across containers. This is a more general formulation of the problem rather than normalizing probabilities across containers which implicitly asserts both that the object must exist and that there is only one to be found---an assumption that may not be valid in the general case.

\subsection{Navigation Cost Distribution} \label{app:cost_distribution}
\begin{figure}
    \centering
    \vspace{0.6em}
    \includegraphics[width=\linewidth]{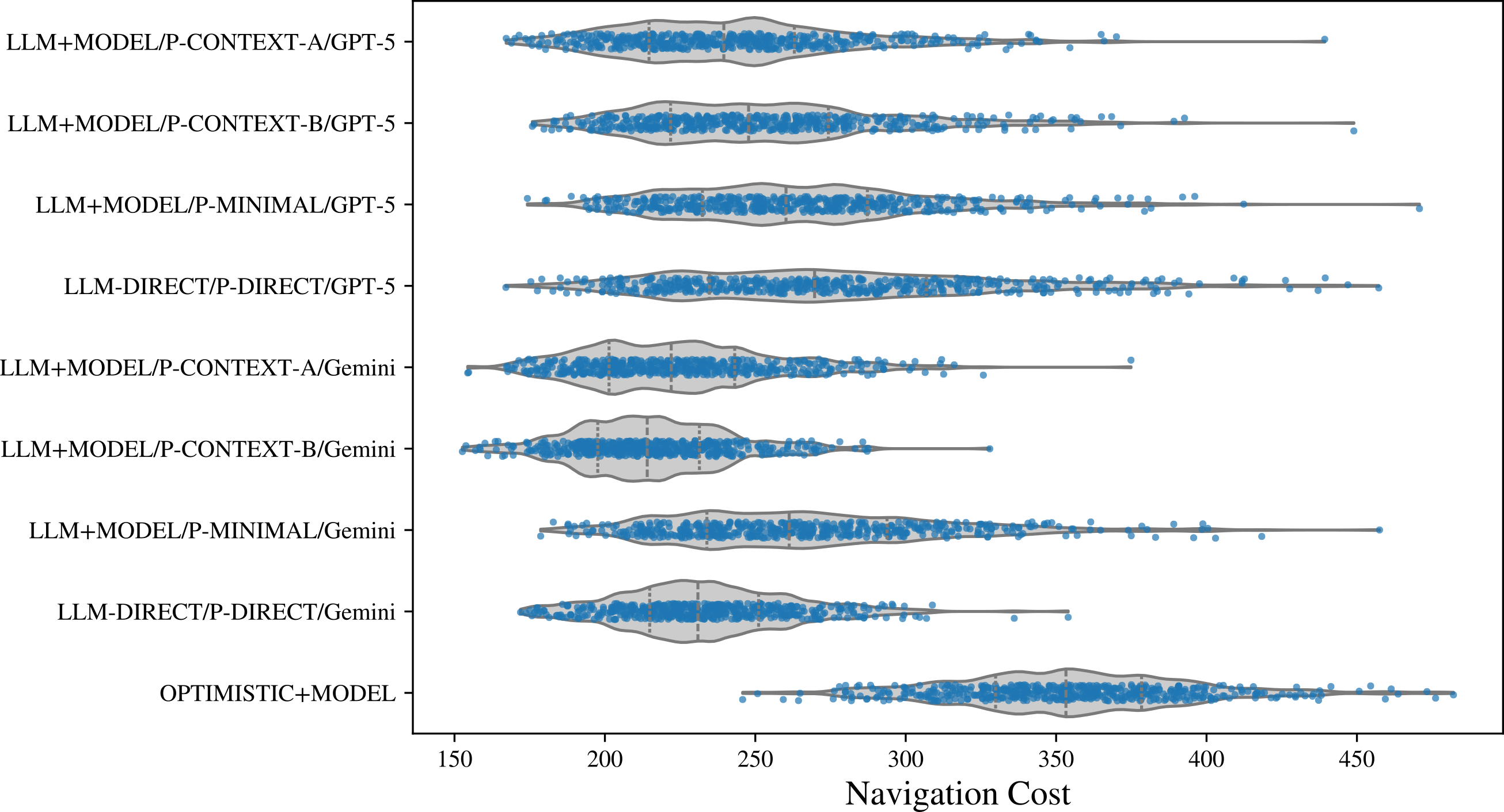}
    \vspace{-2.0em}
    \caption{Distribution of navigation costs for different policy/prompt/\glsxtrshort{LLM} across 150 \gls{ProcTHOR} maps. Dotted lines denote quartiles.}
    \vspace{-0.2em}
    \label{fig:results-costs-violin}
\end{figure}

In \cref{fig:results-costs-violin}, we show violin plots with scatter plots of navigation costs for each policy/prompt/\glsxtrshort{LLM} across 150 \gls{ProcTHOR} maps corresponding to the results in \cref{tab:results-single-prompt} for object search experiments.

\subsection{Object Search Experiments with Open-Source \Glspl{LLM}}
In addition to proprietary models, \gls{GPT}-5 Mini and Gemini 2.5 Flash, we additionally perform object search experiments in 150 \gls{ProcTHOR} maps with two open-source \glspl{LLM}: \gls{GPT-OSS} 120B and Llama3.2 3B. The results are shown in \cref{tab:results-open-source-llms}, which show that our \ourplanner{} planners ourperform baseline \fulllm{} policy that fully relies on \glspl{LLM} for object search, with improvements up to 65.2\% for Llama3.2 and 23.4\% for \gls{GPT-OSS}.

\begin{table}
    \centering
    \footnotesize
    \caption{Navigation costs for object search with open-source \glspl{LLM}}
    \vspace{-0.5em}
    \begin{tabular}{@{}lc@{}}
    \toprule
    \textbf{Policy / Prompt / \Gls{LLM}}     &  \textbf{Avg. Nav. Cost} \\
    \midrule
    \ourplanner{} (ours) / \prompta{} / \gls{GPT-OSS}     & \textbf{217.12} \\
    \ourplanner{} (ours) / \promptb{} / \gls{GPT-OSS}     & 221.47 \\
    \ourplanner{} (ours) / \promptc{} / \gls{GPT-OSS}     & 218.78 \\
    \fulllm{} (baseline) / \promptd{} / \gls{GPT-OSS}     & 283.60 \\
    \midrule
    \ourplanner{} (ours) / \prompta{} / Llama3.2     & 302.83 \\
    \ourplanner{} (ours) / \promptb{} / Llama3.2     & \textbf{281.05} \\
    \ourplanner{} (ours) / \promptc{} / Llama3.2     & 313.24 \\
    \fulllm{} (baseline) / \promptd{} / Llama3.2     & 806.38 \\
    \bottomrule
    \end{tabular}
    \vspace{-2em}
    \label{tab:results-open-source-llms}
\end{table}

\subsection{Token Usage and Costs}
In \cref{tab:token_use_cost}, we show approximate number of tokens used and incurred costs to use proprietary models, \gls{GPT}-5 Mini and Gemini 2.5 Flash, via their respective \glsxtrshortpl{API} for our experiments.
\begin{table}[h]
    \centering
    \footnotesize
    \caption{Token usage and incurred costs for \gls{GPT}-5 and Gemini models}
    \vspace{-0.5em}
    \begin{tabular}{@{}lcccc@{}}
    \toprule
    \multirow{2}{*}{\begin{tabular}[c]{c}\textbf{Model}\\ \textbf{Name}\end{tabular}} & \multicolumn{3}{c}{\textbf{Tokens Used}} & \multirow{2}{*}{\begin{tabular}[c]{c}\textbf{Cost}\\ \textbf{(\textsc{usd})}\end{tabular}} \\ \cmidrule(lr){2-4}
                        & Input    & Output    & Total    &  \\
    \midrule
    GPT-5 Mini          & 903686    & 9264       & 912950    & \$0.14  \\
    Gemini 2.5 Flash    & 904209    &  8066     & 912275    & \$0.24  \\
    \bottomrule
    \end{tabular}
    \label{tab:token_use_cost}
\end{table}

\end{document}